\documentclass{article}
\usepackage{ijcai18}

\usepackage{times}
\usepackage{soul}
\usepackage{url}
\usepackage[hidelinks]{hyperref}
\usepackage[utf8]{inputenc}
\usepackage[small]{caption}
\usepackage{amsmath,amsfonts,amssymb}
\usepackage{color}
\usepackage{graphicx}

\hypersetup{pdfauthor={Tomi Peltola},pdftitle={Local Interpretable Model-agnostic Explanations of Bayesian Predictive Models via Kullback--Leibler Projections}}

\DeclareMathOperator*{\argmin}{arg\,min}

\DeclareMathOperator{\KL}{KL}

\title{Local Interpretable Model-agnostic Explanations of Bayesian Predictive Models via Kullback--Leibler Projections}

\author{
Tomi Peltola$^1$
\\ 
$^1$ Helsinki Institute for Information Technology HIIT, Department of Computer Science,\\ Aalto University, Helsinki, Finland\\
tomi.peltola@aalto.fi
}
\begin{document}

\maketitle

\begin{abstract}
  We introduce a method, KL-LIME, for explaining predictions of Bayesian predictive models by projecting the information in the predictive distribution locally to a simpler, interpretable explanation model. The proposed approach combines the recent Local Interpretable Model-agnostic Explanations (LIME) method with ideas from Bayesian projection predictive variable selection methods. The information theoretic basis helps in navigating the trade-off between explanation fidelity and complexity. We demonstrate the method in explaining MNIST digit classifications made by a Bayesian deep convolutional neural network.
\end{abstract}

\section{Introduction}

Interpreting why a machine learning model makes a certain prediction for given inputs can be difficult for complex models, but is important for evaluating the trustworthiness of the predictions and for model criticism and development. Explaining a prediction requires understanding the proper context of the prediction (inputs and what kind of changes the inputs could meaningfully have) and how the prediction would react to changes in the context (the mapping from inputs to the prediction). Understanding non-linear predictive models globally would require understanding the underlying prediction function at all locations of the input space. Local interpretations are more feasible and many methods have been proposed for this (e.g., \cite{baehrens2010explain,ribeiro2016should,lundberg2017unified}). They rely on the assumption that non-linear functions can be approximated by simpler, more interpretable functions locally.

In addition to interpretability, properly accounting for uncertainty in the predictions is important in many applications, for example, in medicine. In Bayesian probabilistic modelling, the posterior distribution of the model parameters (the conditional distribution induced by the modelling assumptions and training data) and posterior predictive distributions naturally capture the uncertainty about the parameters and model structure (epistemic uncertainty) in addition to noise in prediction (aleatoric uncertainty). When explaining predictions, these uncertainties should also be accounted for.

We introduce a method for Kullback--Leibler divergence based local interpretable model-agnostic explanations, KL-LIME, that extends the recently proposed local interpretation method LIME \cite{ribeiro2016should} to Bayesian models (although it can also be used for non-Bayesian probabilistic models) and provides a principled way to handle different types of predictions (continuous valued, class labels, counts, censored and truncated data, etc.). The proposed method is based on combining the LIME approach with methods introduced for variable selection in Bayesian linear regression models \cite{goutis1998model,dupuis2003variable,peltola2014hierarchical,piironen2017comparison}. The method works by fitting an interpretable explanatory model (e.g., a sparse linear model) to locally match the original prediction model via minimizing the Kullback--Leibler divergence between their predictive distributions.

The next section gives a more detailed background on LIME and Bayesian projection predictive variable selection. Section~\ref{sec:method} describes the proposed method, KL-LIME, and Section~\ref{sec:experiments} demonstrates it in explaining deep convolutional neural network predictions. Section~\ref{sec:discussion} presents a summary and conclusions.

\section{Background}\label{sec:background}

In this section, we describe the LIME method and projection predictive variable selection. In the next section, they are combined to extend LIME to Bayesian predictive models.

\subsection{Local Interpretable Model-agnostic Explanations -- LIME}

The Local Interpretable Model-agnostic Explanations (LIME) method of Ribeiro et al.\ \shortcite{ribeiro2016should} provides local explanations of predictions of a classifier $f$ by fitting a simpler, interpretable explanation model $g$ locally around the data point $x$ of which classification is to be explained. The explanation model is fit on an interpretable representation of the original data space. For example, let $x \in \mathbb{R}^d$ be a vector of the gray scale values of pixels in an image. An interpretable representation $x' \in \{0,1\}^{d'}$ might then be a vector of binary values representing the absence or presence of pixels in the image (absence meaning having the value of a background color, e.g., white). The LIME explanation $\hat{g}$ arises by solving the optimization problem
\begin{equation*}
  \hat{g} = \argmin_{g \in G} \mathcal{L}(f, g, \pi_x) + \Omega(g),
\end{equation*}
where $G$ is the explanation model family, $\mathcal{L}$ is a loss function, $\pi_x$ defines the locality around $x$, and $\Omega$ is a complexity penalty.

In practice, $G$ is taken to be the set of linear regression models, with $\Omega$ restricting that only some number of the explanatory features can have non-zero regression weights (although other types of explanation models could be used). The loss function is taken to be the weighted L2 distance
\begin{equation*}
  \mathcal{L}(f, g, \pi_x) = \sum_i \pi_x(z_i) (f(z_i) - g(z_i'))^2,
\end{equation*}
where the sum goes over a set of sampled perturbed points around $x$, $\{(z_i, z_i'), i = 1,\ldots,m\}$, where $z_i$ is a perturbed data point in the original data space and $z_i'$ the corresponding interpretable representation. $\pi_x(z_i)$ weights the samples based on their similarity to $x$, the point where the classification result is being explained.

\subsection{Projection predictive variable selection} 

Variable selection is the problem of choosing a smaller set of covariates from among the full set available and is often used to simplify high-dimensional regression models. In Bayesian modelling, sparsity-inducing and shrinkage priors can be used to regularize high-dimensional regression models. However, they do not lead to truly sparse posterior distributions, since, with finite data, there will remain uncertainty about whether some covariates should or should not be included in the regression. Projection predictive variable selection is an approach to variable selection in Bayesian regression models, which removes covariates that do not considerably contribute to the explanatory power of the full model \cite{goutis1998model,dupuis2003variable,peltola2014hierarchical,piironen2017comparison}. It works by projecting the information in the model encompassing all variables to a model that uses only a subset of the variables and has been empirically shown to have competitive performance compared to other variable selection approaches \cite{piironen2017comparison}.

For a prediction model $M$, defined on the full set of covariates, let $p(y \mid x, \theta, M)$ be the observation model of the target variable $y$ given covariates $x$ and model parameters $\theta$ and $p(\theta \mid D, M)$ the posterior distribution of the parameters given a training dataset $D$. Given the parameter $\theta$ of the full model $M$, the projection predictive variable selection approach fits a model $M_s$ with a subset $s$ of the covariates, denoted by $x_s$ with $x = (x_s, x_{\backslash s})$, by minimizing the Kullback--Leibler divergence to the full model $M$
\begin{equation*}
  \hat{\theta}_s = \argmin_{\theta_s} \frac{1}{N} \sum_{i=1}^N  \KL[p(y \mid x_i, \theta, M) \| p(y \mid x_{i,s}, \theta_s, M_s)],
\end{equation*}
where $i$ runs over the $N$ training samples in the dataset $D$. In practice, the optimization is solved $L$ times for $L$ samples $\theta^{(l)}, l = 1,\ldots,L$, from the posterior distribution $p(\theta \mid D, M)$ to get a projected posterior distribution for the model $M_s$. The total information loss $\delta$ of using the subset $x_s$ instead of full set $x$ is then approximated as the average of the above loss over the samples from the posterior distribution:
\begin{equation*}
  \begin{aligned}
    & \delta[M \| M_s] \\ &\approx \frac{1}{L N} \sum_{l=1}^L \sum_{i=1}^N \KL[p(y \mid x_i, \theta^{(l)}, M) \| p(y \mid x_{i,s}, \hat{\theta}^{(l)}_s, M_s)].
  \end{aligned}
\end{equation*}
For generalized linear models, the optimization problems are related to the generalized linear model estimation equations \cite{goutis1998model}. The projection method is, however, not limited to linear regression. For example, the approach has also been used for variable selection in Gaussian processes \cite{piironen2016projection}.  A similar KL divergence minimization approach, with a further regularization penalty, is used in \cite{tran2012predictive} to define a predictive lasso method.

For variable selection, the projections are supplanted with a search process for finding a good subset of the covariates. This needs to weigh the benefits and costs of keeping a number of covariates. Dupuis and Robert \shortcite{dupuis2003variable} introduced the relative explanatory power 
\begin{equation}
  1 - \frac{\delta[M \| M_s]}{\delta[M \| M_0]},\label{eqn:exp_power}
\end{equation}
where $M_0$ is a null model (e.g., the model without any covariates), to quantify the quality of the subset models and to help in determining the best subset model by choosing the smallest model that retains enough of the explanatory power of the full model. An alternative approach was suggested in \cite{peltola2014hierarchical}, where cross-validation was combined with the projection predictive approach to estimate out-of-sample prediction performances at each model size along a forward selection path. 

\section{KL-LIME for explaining predictions of Bayesian models}\label{sec:method}

We combine ideas from LIME and projection predictive variable selection to define the KL-LIME method for Kullback--Leibler divergence based local interpretable model-agnostic explanations of Bayesian predictive models (although the method can also be applied on non-Bayesian probabilistic models). Let $p(y \mid x, \theta, M)$ be the observation model of the predictive model $M$ given input $x$ and $p(\theta \mid D, M)$ the posterior distribution of its parameters $\theta$ given a dataset $D$. Similar to LIME, we define the explanation as an interpretable model $p(y \mid x', \phi, G)$ from (now a probabilistic) model family $G$ with parameters $\phi$, and possibly operating on a simplified representation $x'$ of the original input $x$. Similar to projection predictive variable selection, the parameters of the explanation model are found by minimizing its Kullback--Leibler divergence from the predictive model $M$:
\begin{equation*}
  \begin{aligned}
    \hat{\phi}^{(l)} &= \argmin_\phi && \!\!\!\!\int \pi_x(z)  \KL[p(y \mid z, \theta^{(l)}, M) \| p(y \mid z', \phi, G)] dz \\ & && \!\!\!\!+ \Omega(\phi)\\
                     &\approx \argmin_\phi && \!\!\!\!\frac{1}{N} \sum_{i=1}^N \KL[p(y \mid z_i, \theta^{(l)}, M) \| p(y \mid z_i', \phi, G)] \\ & && \!\!\!\!+ \Omega(\phi),
  \end{aligned}
\end{equation*}
for $l=1,\ldots,L$ posterior samples from $p(\theta \mid D, M)$ and where $\pi_x(z)$ is a probability distribution (we assume there is a mapping between $z$ to $z'$) defining the local input data space neighborhood around $x$, the data point which prediction is to be explained. $\Omega$ penalizes the complexity of the explanation model. In practice, the expectation over locality $\pi_x$ is computed via a Monte Carlo approximation by sampling $N$ points from $\pi_x(z)$.

To measure the fidelity of the explanation, we can compute the relative explanatory power between $M$ and $G$, following Equation~\ref{eqn:exp_power}, with an appropriate definition of the null model. Alternatively, one could also compute more direct measures of the performance, such as mean squared error or classification accuracy by sampling a test set from the locality distribution $\pi_x(z)$.

In the demonstration of the approach below, we use linear models for the explanation model family and L1 regularization for the complexity penalty $\Omega$. The KL minimization can then be solved with lasso regression (or generalized linear model variants of lasso regression). For non-Bayesian probabilistic models, one would not have posterior samples, but only a point estimate of the parameters $\theta$ which can be projected to the explanation parameters $\phi$.

\section{Demonstration}\label{sec:experiments}

We demonstrate the proposed method in explaining deep convolutional neural network predictions in the MNIST dataset of images of digits, with the task of classifying between 3s and 8s. The neural network has two convolutional layers and two fully connected layers and uses ReLU activation functions\footnote{A slightly customized version of the PyTorch MNIST example is used, \url{https://github.com/pytorch/examples/tree/master/mnist/}, accessed on May 17th, 2018. This achieves about 99.2\% accuracy on the test data.}. Bayesian inference is approximated with the Bernoulli dropout method \cite{gal2016dropout,gal2016Bayesian}, with a dropout probability of 0.2 and 100 Monte Carlo samples at test time, which provides a rough approximation of the model uncertainty for prediction. The locality distribution around image $x$ is defined by randomly zeroing pixels (i.e., setting them to the background value of white color) by first sampling the zeroing probability from a beta distribution and then sampling a binary mask with this probability in independent Bernoulli distributions. The simplified representation has 1 for pixels that are not at the background value and 0 otherwise. 1,000 samples are drawn from the locality distribution and used as data for fitting the explanation models with KL-LIME. The explanation model is a linear logistic regression model with L1 penalty.

\begin{figure*}
  \includegraphics[width=\textwidth]{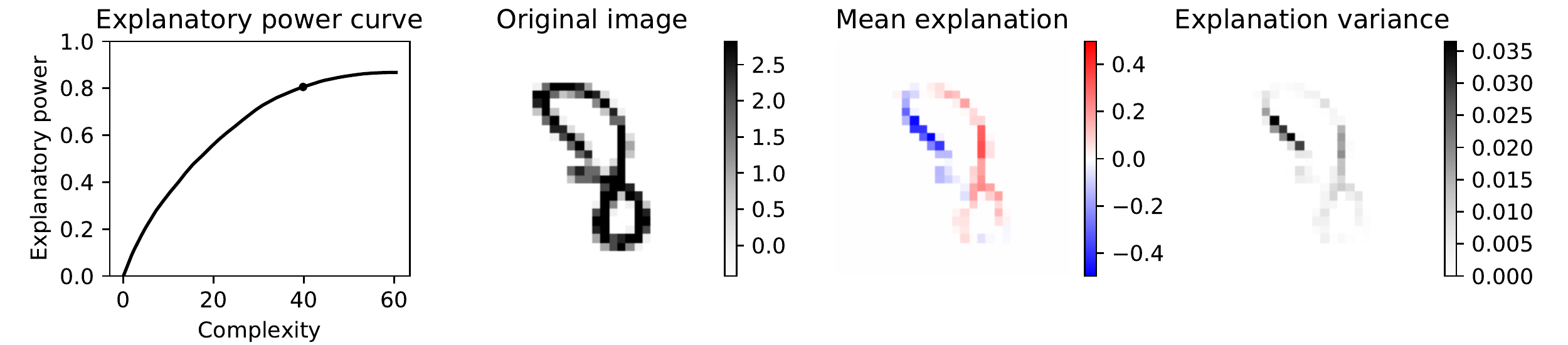}\\

  \includegraphics[width=\textwidth]{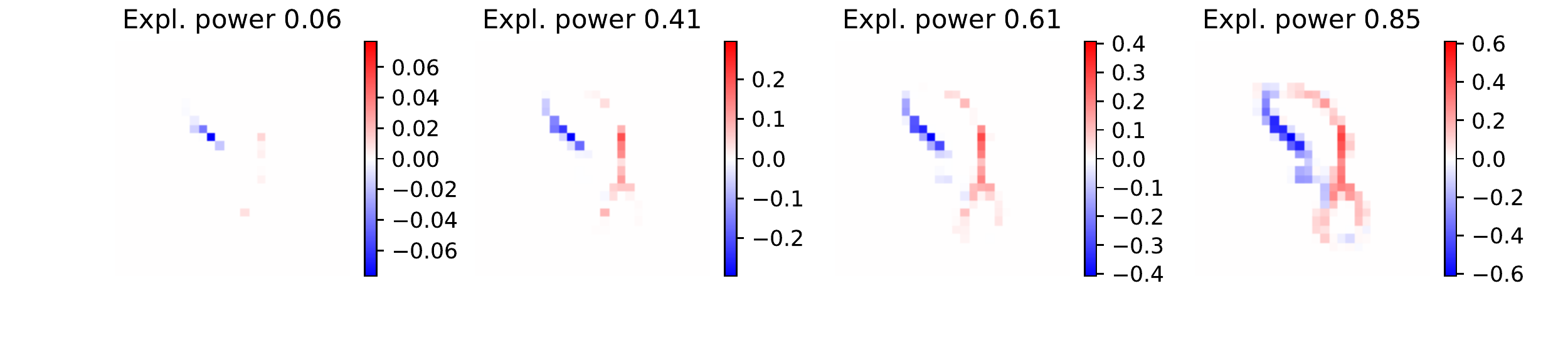}\\

  \vspace{-10mm}

  \includegraphics[width=\textwidth]{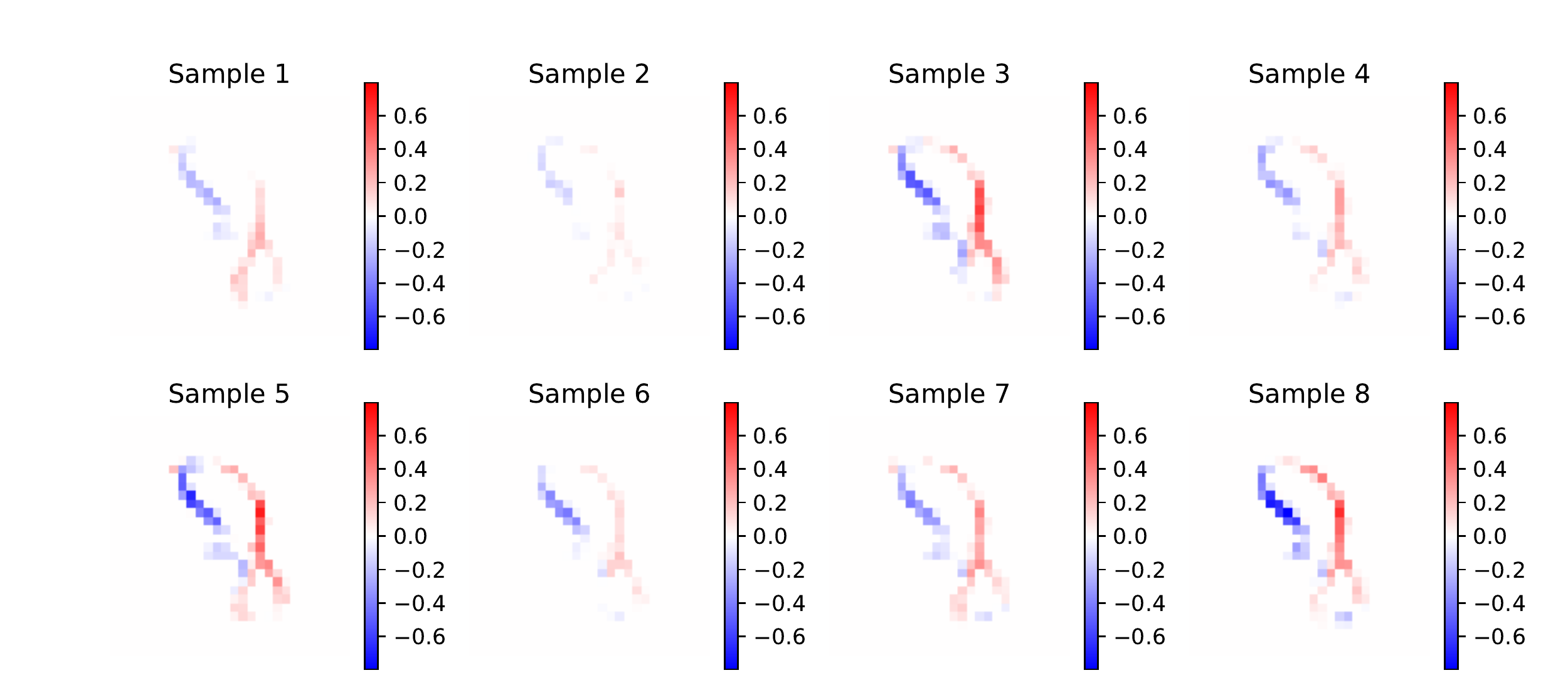}
  \caption{Example explanation of an image of an 8 that was misclassified as 3. Top row: Explanatory power curve shows the relative explanatory power as a function of the explanation complexity (here, the mean number of active pixels in the explanation). Dot shows the complexity chosen for the shown explanation (expl.\ power of 0.8). Mean explanation gives the posterior mean of the projected explanations and variance gives the uncertainty in the explanation. Second row: Explanations with different complexities and relative explanatory powers. Bottom two rows: Individual posterior samples of explanations.}\label{fig:exp1}
\end{figure*}

The top row of Figure~\ref{fig:exp1} shows an example of explaining an image of 8 that is misclassified by the classifier. The relative explanatory power curve can be used to determine the trade-off between explanation fidelity and complexity. In this case, the curve plateaus around 0.85 explanatory power, showing that it's not possible to attain perfect fidelity with the chosen explanation model. The mean explanation shows the posterior mean of the projected parameters of the explanation model. The pixels in the left hand side of the upper loop of the 8 steer the classification towards an 8 as expected for classifying between 8s and 3s. However, the explanation implies that the model has considered the right hand side parts of the image as pointing the classification to a 3. The variance of the explanation is largest in the left hand side of the upper loop of the digit.

The second row of Figure~\ref{fig:exp1} shows mean explanations at different trade-offs between the fidelity and complexity. The lowest level of explanatory power does not capture the classification particularly well. In the case of images as here, even the most complex explanations (by the measure of how many pixels are active in the linear explanation model) are often readily interpretable, since humans are good at image perception. In many other cases, such as textual data or quantitative covariates, say, in personalized medicine, a proper trade-off between complexity and fidelity would be more important. Finally, the two bottom rows of Figure~\ref{fig:exp1} show individual posterior samples of the explanation model. This allows getting a more complete picture of the model uncertainty reflected in the explanations.

\begin{figure*}
  \includegraphics[width=\textwidth]{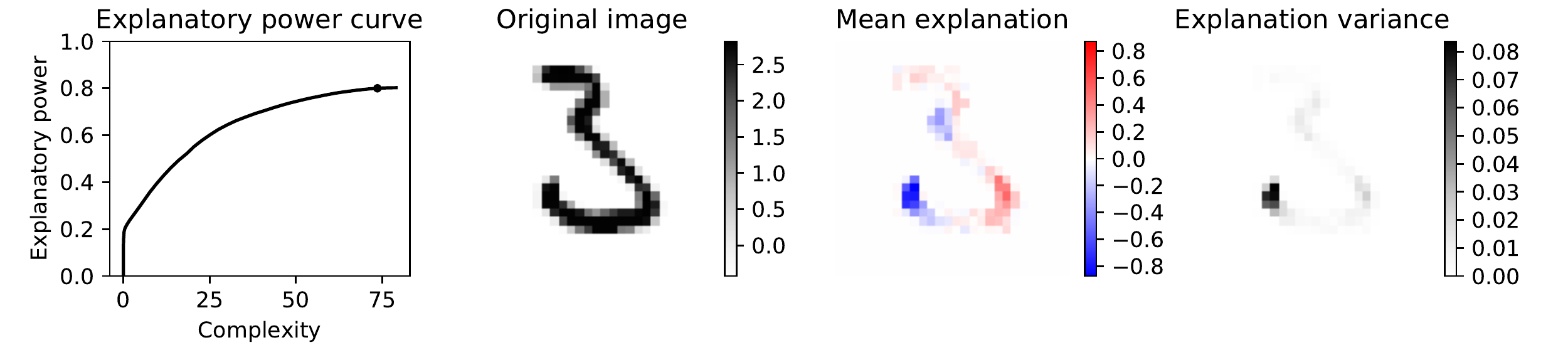}
  \caption{Example explanation of an image of a 3 that was misclassified as 8. See Figure~\ref{fig:exp1} caption for the description of the panels.}\label{fig:exp2}
\end{figure*}

Figure~\ref{fig:exp2} shows another example of an explanation: in this case of an image of a 3 that is misclassified as an 8. The explanation implies that this happens because of the elongated lower left curve in the digit.

\section{Conclusion}\label{sec:discussion}

We presented a method, KL-LIME, to construct local interpretable explanations of Bayesian predictive models by projecting the information in the predictive distribution of the model to a simpler, interpretable probabilistic explanation model. The approach is based on combining ideas from the recent Local Interpretable Model-agnostic Explanations (LIME) method \cite{ribeiro2016should} and Bayesian projection predictive variable selection. This allows accounting for model uncertainty also in the explanations.

The approach gives a principled way of extending LIME to different types of predictions and explanations as long as we can compute the Kullback--Leibler divergence between the predictive distributions for the original model and the explanation model and minimize it to fit the explanation model. In particular, within this constraint, both the original task and the explanation model can be arbitrarily changed without losing the information theoretical interpretation of the projection for finding the explanation model. The value of the KL divergence can be used as a measure of the fidelity of the explanation. 

\appendix

\bibliographystyle{named}
\bibliography{whi}

\end{document}